\documentclass[sigconf]{acmart}

\AtBeginDocument{%
  }

\usepackage{enumitem}
\setlist{nosep}
\usepackage{mdframed}
\mdfdefinestyle{promptbox}{
  backgroundcolor=gray!10,
  linecolor=black,
  linewidth=0.5pt,
  roundcorner=5pt,
  innertopmargin=6pt,
  innerbottommargin=6pt,
  innerleftmargin=8pt,
  innerrightmargin=8pt
}
\usepackage[most]{tcolorbox}
\tcbuselibrary{listingsutf8}

\copyrightyear{2025}
\acmYear{2025}
\setcopyright{acmlicensed}\acmConference[MM '25]{Proceedings of the 33rd ACM International Conference on Multimedia}{October 27--31, 2025}{Dublin, Ireland}
\acmBooktitle{Proceedings of the 33rd ACM International Conference on Multimedia (MM '25), October 27--31, 2025, Dublin, Ireland}
\acmDOI{10.1145/3746027.3762039}
\acmISBN{979-8-4007-2035-2/2025/10}

\usepackage{pifont}
\newcommand{\cmark}{\ding{51}}%

\usepackage{pgfplots}
\usepackage{multirow} 
\usepackage{balance}
\pgfplotsset{compat=1.18}

\begin{document}

\title[ReCap: Event-Aware Image Captioning with Article Retrieval and Semantic Gaussian Normalization]{ReCap: Event-Aware Image Captioning with Article Retrieval and Semantic Gaussian Normalization}



\author{Thinh-Phuc Nguyen}
\orcid{0009-0000-5150-9307}
\affiliation{%
  \institution{University of Science, VNU-HCM}
  \city{Ho Chi Minh}
  \country{Vietnam}
}

\author{Thanh-Hai Nguyen}
\orcid{0009-0004-7953-8942}
\affiliation{%
  \institution{University of Science, VNU-HCM}
  \city{Ho Chi Minh}
  \country{Vietnam}
}

\author{Gia-Huy Dinh}
\orcid{0009-0004-8746-3004}
\affiliation{%
  \institution{University of Science, VNU-HCM}
  \city{Ho Chi Minh}
  \country{Vietnam}
}

\author{Lam-Huy Nguyen}
\orcid{0009-0003-2890-5741}
\affiliation{%
  \institution{University of Science, VNU-HCM}
  \city{Ho Chi Minh}
  \country{Vietnam}
}

\author{Minh-Triet Tran}
\orcid{0000-0003-3046-3041}
\affiliation{%
  \institution{University of Science, VNU-HCM}
  \city{Ho Chi Minh}
  \country{Vietnam}
}

\author{Trung-Nghia Le}
\orcid{0000-0002-7363-2610}
\affiliation{%
  \institution{University of Science, VNU-HCM}
  \city{Ho Chi Minh}
  \country{Vietnam}
}
\authornote{Corresponding author. Email: ltnghia@fit.hcmus.edu.vn}

\renewcommand{\shortauthors}{Thinh-Phuc Nguyen, Thanh-Hai Nguyen, Gia-Huy Dinh, Lam-Huy Nguyen, Minh-Triet Tran, and Trung-Nghia Le}

\begin{abstract}
Image captioning systems often produce generic descriptions that fail to capture event-level semantics which are crucial for applications like news reporting and digital archiving. We present ReCap, a novel pipeline for event-enriched image retrieval and captioning that incorporates broader contextual information from relevant articles to generate narrative-rich, factually grounded captions. Our approach addresses the limitations of standard vision-language models that typically focus on visible content while missing temporal, social, and historical contexts. ReCap comprises three integrated components: (1) a robust two-stage article retrieval system using DINOv2 embeddings with global feature similarity for initial candidate selection followed by patch-level mutual nearest neighbor similarity re-ranking; (2) a context extraction framework that synthesizes information from article summaries, generic captions, and original source metadata; and (3) a large language model-based caption generation system with Semantic Gaussian Normalization to enhance fluency and relevance. Evaluated on the OpenEvents V1 dataset as part of Track 1 in the EVENTA 2025 Grand Challenge, ReCap achieved a strong overall score of 0.54666, ranking 2nd on the private test set. These results highlight ReCap’s effectiveness in bridging visual perception with real-world knowledge, offering a practical solution for context-aware image understanding in high-stakes domains. The code is available at \url{https://github.com/Noridom1/EVENTA2025-Event-Enriched-Image-Captioning}.
\end{abstract}

\ccsdesc[500]{Information systems~Information retrieval}
\ccsdesc[500]{Computing methodologies~Computer vision}
\ccsdesc[500]{General and reference~Metrics}

\keywords{image captioning, image retrieval, real-world events, event-centric vision-language}
\begin{teaserfigure}
      \includegraphics[width=\textwidth]{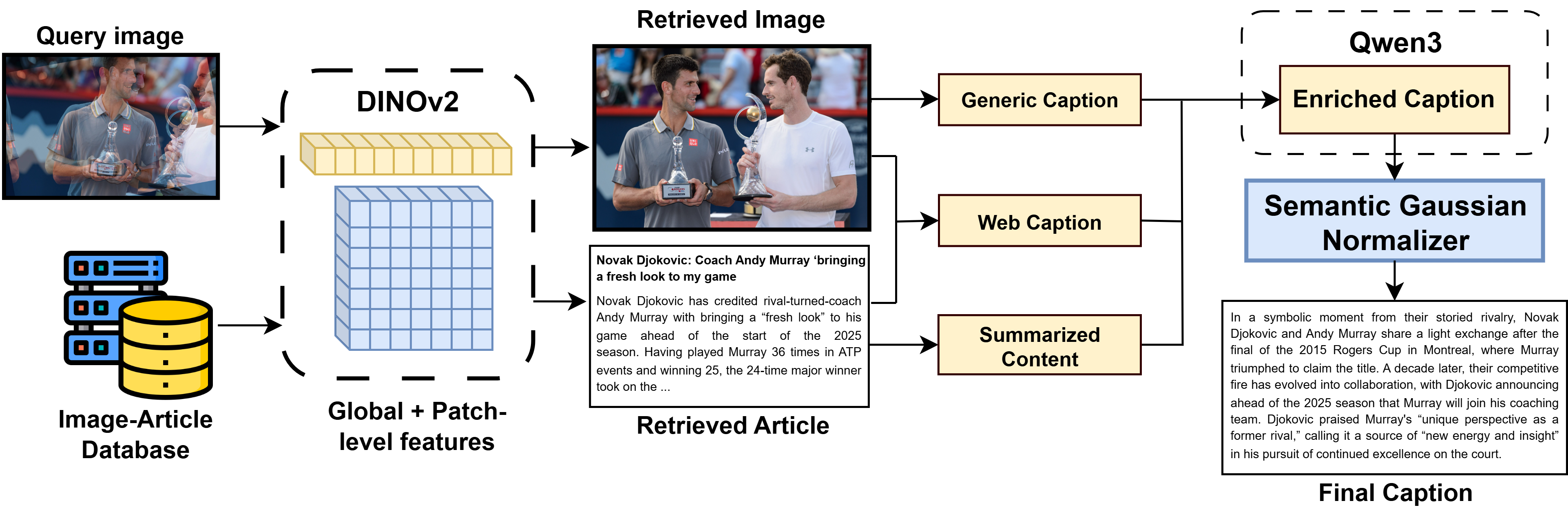}
    \vspace{-8mm}
    \Description{Overall pipeline of the proposed ReCap, showing the process from query image to enriched caption.}
    \caption{Overall pipeline of the proposed ReCap. A heavily transformed query image is processed by DINOv2 to extract both global and patch-level visual features. These are used to retrieve the corresponding image from a database. The retrieved image and its associated article are used to extract multiple contextual signals, including a generic visual caption, a web-scraped caption, and a summary of the article content. These inputs are fused via the Qwen3 language model to generate an enriched caption, which is then refined by a Semantic Gaussian Normalization module to produce the final output.}
  \label{fig:teaser}
\end{teaserfigure}


\maketitle

\section{Introduction}
\label{sec1: Introduction}

In real-world applications like news reporting and digital archiving, images often hold meaning beyond visual content—they may reflect historical, political, or social events. Integrating such context into captions enhances informativeness, factual grounding, and semantic depth. However, conventional captioning models often produce generic descriptions, lacking the event-level semantics crucial for domains like journalism and documentation~\cite{geo-aware, beyond-generic, informative-external}.

While vision-language models (VLMs) describe what is seen, they rarely explain why it happens or how it connects to real-world narratives~\cite{visual-news, transform-tell-entity}. For example, a caption like “a group of people standing outside” omits key details—who they are, what event they’re part of, and why they’re there.

This gap has critical consequences in high-stakes settings such as disaster response or political journalism, where vague captions risk misinformation or archival loss. To address this, the task of \textit{event-enriched image retrieval and captioning} aims to generate context-aware captions by combining visual features with external textual sources, such as associated articles. Prior work has explored document-level cues~\cite{goodnews, nytimes800k}, OCR-based scene text~\cite{m4c, textcaps}, structured knowledge graphs~\cite{boosting-entity, transform-tell-entity}, and commonsense reasoning~\cite{visualcomet}. Named entities, in particular, have shown promise in anchoring captions to specific people, places, and events~\cite{goodnews, report-scenario}.

In alignment with these developments, the EVENTA 2025 Grand Challenge~\cite{eventa25} introduces Track 1: \textit{Event-Enriched Image Retrieval and Captioning}\footnote{\url{https://ltnghia.github.io/eventa/eventa-2025/track1}}, which aims to push the boundaries of context-grounded image understanding. Participants are tasked with retrieving the correct article associated with a heavily transformed query image and generating a caption that reflects not just visual elements but also the underlying event, temporal, and social context. 

In response to this challenge, we propose \textit{ReCap}, a retrieval-augmented captioning pipe\-line designed for this challenge, which comprises three main stages: Article Retrieval, Event-enriched Image Captioning, and Semantic Gaussian Normalization. Given a query image, we extract its global features using DINOv2~\cite{oquab2023dinov2} and identify the top-100 most visually similar images in the database via cosine similarity. To improve retrieval precision, we re-rank the candidates using patch-level features with a mutual nearest neighbor-based aggregation strategy. Once the correct article is retrieved, we combine the article content and the retrieved image with outputs from a Qwen-based vision-language model and a Qwen-based large language model to produce a rich, context-aware caption. After enrichment, the generated caption undergoes further refinement via Semantic Gaussian Normalizer, a module specifically tailored for agreement with ground truth captions. The source code is publicly available\footnote{\url{https://github.com/Noridom1/EVENTA2025-Event-Enriched-Image-Captioning}}.

The EVENTA 2025 Grand Challenge dataset \cite{openeventsv1} contains more than 200,000 news articles from two well-known sources-CNN and The Guardian-spanning various topics, which provide background context for caption enrichment tasks. The test set for Track 1 features heavily transformed query images, presenting a significant challenge for both retrieval and captioning systems. On the private test set, ReCap achieved the second-highest overall score of 0.54666 (mAP: 0.982, R@1: 0.977, R@10: 0.988, CLIP Score: 0.870, CIDEr: 0.205), demonstrating strong performance in both retrieval and captioning. 

Our contributions are as follows:
\begin{itemize}
    \item We introduce a two-stage retrieval strategy combining global feature-based filtering with fine-grained patch-level re-ranking via mutual nearest neighbor matching.
    \item We develop a hybrid captioning framework that fuses visual-language outputs, article summaries, and LLM-based reasoning to produce event-aware captions grounded in real-world context.
    \item Our approach achieves second place on the EVENTA 2025 Challenge Track 1, validating the effectiveness of our method in large-scale retrieval and event-level captioning.
\end{itemize}

\section{Related Work}
\label{sec2: Related Work}


\subsection{Image Retrieval}

\textbf{Feature Extraction Methods.} Self-supervised learning has transformed feature extraction for image retrieval. DINOv2 \cite{oquab2023dinov2} extracts rich visual features without labeled data using a Vision Transformer architecture trained on a curated dataset. Its ability to capture both global and local semantic information makes it valuable for challenging retrieval scenarios with transformed query images \cite{oquab2023dinov2, jose2025dinov2}.

\textbf{Global Feature-based Retrieval.} Image retrieval systems typically use global features for initial retrieval and local features for reranking \cite{shao2023global}. DINOv2's CLS token embeddings serve as effective global descriptors, capturing semantic content while maintaining robustness to transformations. Shao et al. \cite{shao2023global} demonstrated that well-optimized global features can achieve competitive performance on standard retrieval benchmarks.

\subsection{Reranking Strategies}

\textbf{Patch-level Matching and Mutual Nearest Neighbors.} Mutual k-nearest neighbor (k-nn) graphs \cite{li2015image} effectively distinguish correct matches from incorrect ones by constructing relationships where candidates receive scores only from their mutual k nearest neighbors. Patch-level matching treats retrieval as a local correspondence problem, quantifying similarity through patch feature overlap \cite{shivakumara2011novel}. This approach excels with occluded or transformed images.

\textbf{Efficient Reranking Algorithms.} Recent work has developed lightweight reranking approaches incorporating both local and global features. Reranking Transformers \cite{tan2021instance} process top matching results in a single forward pass. Other approaches reformulate reranking as Graph Neural Networks \cite{zhang2020understanding}, separating retrieval and feature updating phases. Approximate nearest neighbor techniques offer various trade-offs between accuracy and efficiency \cite{hambarde2023information}.

Our work builds on these advances to develop a two-stage retrieval system using DINOv2 embeddings with global feature similarity for initial selection followed by patch-level mutual nearest neighbor reranking.


\subsection{Event-Enriched Image Captioning}

Traditional image captioning methods typically follow an encoder-decoder architecture (e.g., CNN + LSTM/Transformer) to generate general visual descriptions~\cite{zeng2024meacap, deepcap, yang2024samt, image_semantic, hossain2019comprehensive, vinyals2016show, aneja2018convolutional, herdade2019image}. While effective at capturing objects and scenes, they often lack event-level semantics and contextual grounding.

To address this, entity-aware methods integrate external knowledge. Tran et al. align image regions with entity mentions via multi-modal attention~\cite{transform-tell-entity}. Zhao et al. build multi-modal knowledge graphs to link image objects and textual entities~\cite{boosting-entity}, while Hu et al. propose a progressive model that retrieves and attends to relevant sentences for entity-rich captions~\cite{hu2020icecap}.

Other approaches incorporate broader text signals. Biten et al. leverage article headlines and bodies~\cite{goodnews}; M4C\cite{m4c} and TextCaps~\cite{textcaps} use OCR to integrate scene text into caption generating process.

Commonsense and narrative reasoning are also explored. VisualCOMET~\cite{visualcomet} predicts past and future events from images, while rule-driven models proposed by Xu et al. guide captioning using structured templates~\cite{rule-driven}.

News-oriented models combine retrieval and captioning. Visual News Captioner~\cite{visual-news} uses article metadata to predict entities; VACNIC~\cite{qu2023visually} retrieves contextual sentences with CLIP and links faces to names; and multi-modal entity prompting~\cite{zhang2022fine} enhances captioning with relevant entities.

Despite progress, most methods assume aligned text is readily available and struggle with retrieving the correct article for transformed or ambiguous images. Our framework addresses this by first retrieving relevant articles using both global and patch-level visual features, then generating enriched captions by fusing visual content, article summaries, and web-sourced descriptions. We also introduce a Semantic Gaussian Normalizer to adaptively control caption length and informativeness, producing event-aware, contextually grounded outputs.

\section{Proposed Method}
\label{sec3: Proposed Method}

\subsection{Overview}

We propose a novel pipeline for event-enriched image retrieval and captioning that combines visual features and real-world textual context to generate narrative-rich, context-aware captions. The task requires not only recognizing image content but also understanding the underlying event semantics.

As illustrated in \autoref{fig:teaser}, our method comprises three main stages: (1) Article Retrieval, (2) Context Extraction, and (3) Caption Generation with Semantic Gaussian Normalization. A transformed query image is first matched to its corresponding article using a two-stage retrieval strategy based on DINOv2 features—global similarity for initial filtering and patch-level bi-directional matching for reranking. From the retrieved article, we extract contextual cues including the article summary, a generic caption, and a scraped caption from the source URL. These signals are fused into a prompt for a large language model to generate an enriched caption. Finally, we apply our Semantic Gaussian Normalizer to adjust length and semantics, enhancing fluency and alignment with evaluation metrics, particularly CIDEr.

\subsection{Article Retrieval}

\begin{figure}[t!]
  \includegraphics[width=\linewidth]{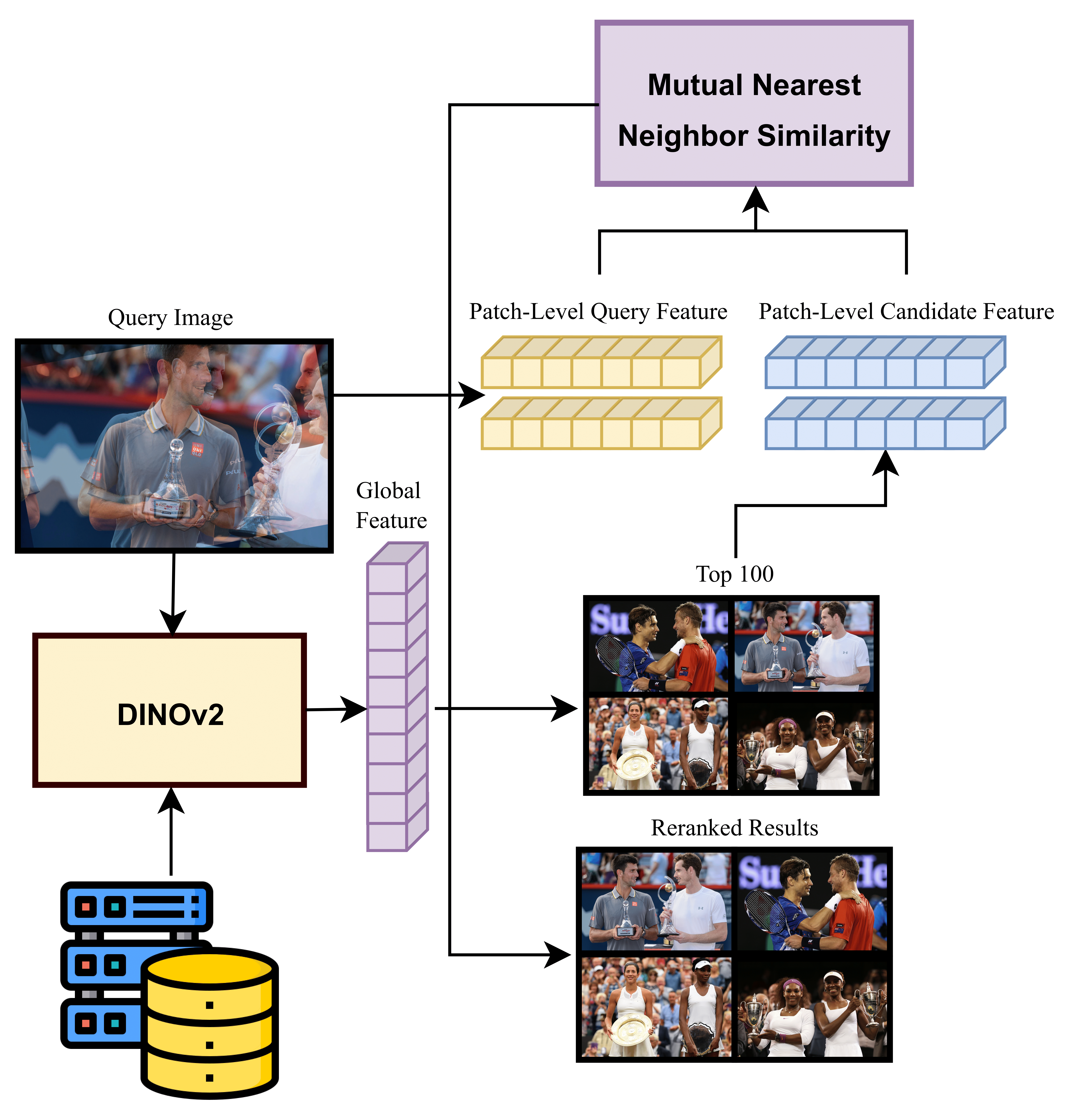}
  \vspace{-5mm}
  \caption{Pipeline for Article and Image Retrieval and Re-ranking using Patch-level Mutual Nearest Neighbor Similarity}
  \label{fig:retrieveFig}
  \vspace{-5mm}
\end{figure}

Accurately retrieving the relevant article for a given image is critical for generating event-enriched captions. The process shown in \autoref{fig:retrieveFig} is divided into two stages: Selecting candidate images via global feature and re-ranking using patch-level mutual nearest neighbor similarity

\textbf{a) Global Feature Similarity for Candidate Selection:} We use DINOv2 \cite{oquab2023dinov2} (ViT-g/14	with registers), a state-of-the-art without supervision model, to extract global image embeddings. These embeddings are compared with entries in the article database using cosine similarity to retrieve an initial set of top-100 candidate articles. The global features capture high-level semantic information, making this stage efficient for coarse filtering.

\textbf{b) Patch-Level Mutual Nearest Neighbor Similarity (MMNS) Re-ranking Strategy:} To improve retrieval recall explicitly R@1 and R@10, for each query image and its corresponding top-100 candidates from Stage 1, we re-extract dense local descriptors. Specifically, we use the same DINOv2 \cite{oquab2023dinov2} model to produce per-patch embeddings by discarding the classification token and retain the remaining patch tokens. 

Let \(P\) denote the number of patches in an image and \(D\) the embedding dimension. For a given query image, we represent its patch embeddings as a matrix \(Q \in \mathbb{R}^{P \times D}\), while for a candidate image we use the matrix \(C \in \mathbb{R}^{P \times D}\). 

\begin{figure}[t!]
  \includegraphics[width=\linewidth]{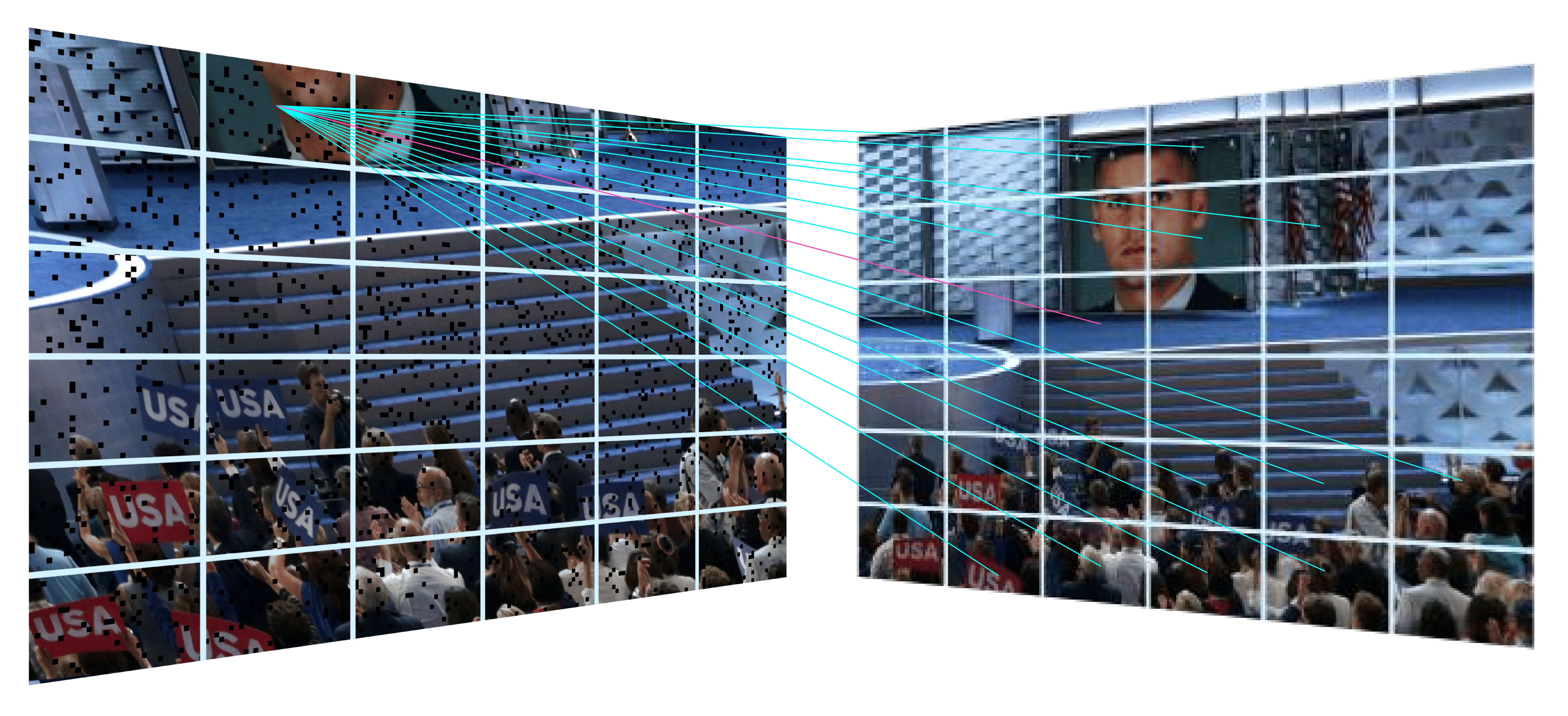}
  \vspace{-5mm}
  \caption{Visualization Patch-level Mutual Nearest Neighbor Similarity}
  \label{fig:patch_visual}
  \vspace{-5mm}
\end{figure}

To assess the similarity between the query and each candidate, we perform patch-level mutual nearest neighbor matching. In the forward direction, for each patch in the query image, we compute the maximum cosine similarity across all candidate patches as shown in \autoref{fig:patch_visual}. In the backward direction, we similarly find, for each candidate patch, its highest cosine similarity to any query patch.

These mutual maximum nearest neighbor similarity (MMNS) scores are then aggregated and used to re-rank the top 100 candidates identified in the initial retrieval stage to efficiently boosts the retrieval performance.
\begin{equation}
    MNNS(Q, C) = \frac{1}{2} \Bigg( \frac{1}{P_Q} \sum_{q_i \in Q} \max_{c_j \in C} \text{sim}(q_i, c_j) 
+ \frac{1}{P_C} \sum_{c_j \in C} \max_{q_i \in Q} \text{sim}(c_j, q_i) \Bigg).
\end{equation}


This two-stage approach ensures robust retrieval by combining the efficiency of global feature matching with the precision of patch-level reranking.

\subsection{Event-enriched Image Captioning}

\begin{figure}[t!]
    \centering
    \includegraphics[width=\linewidth]{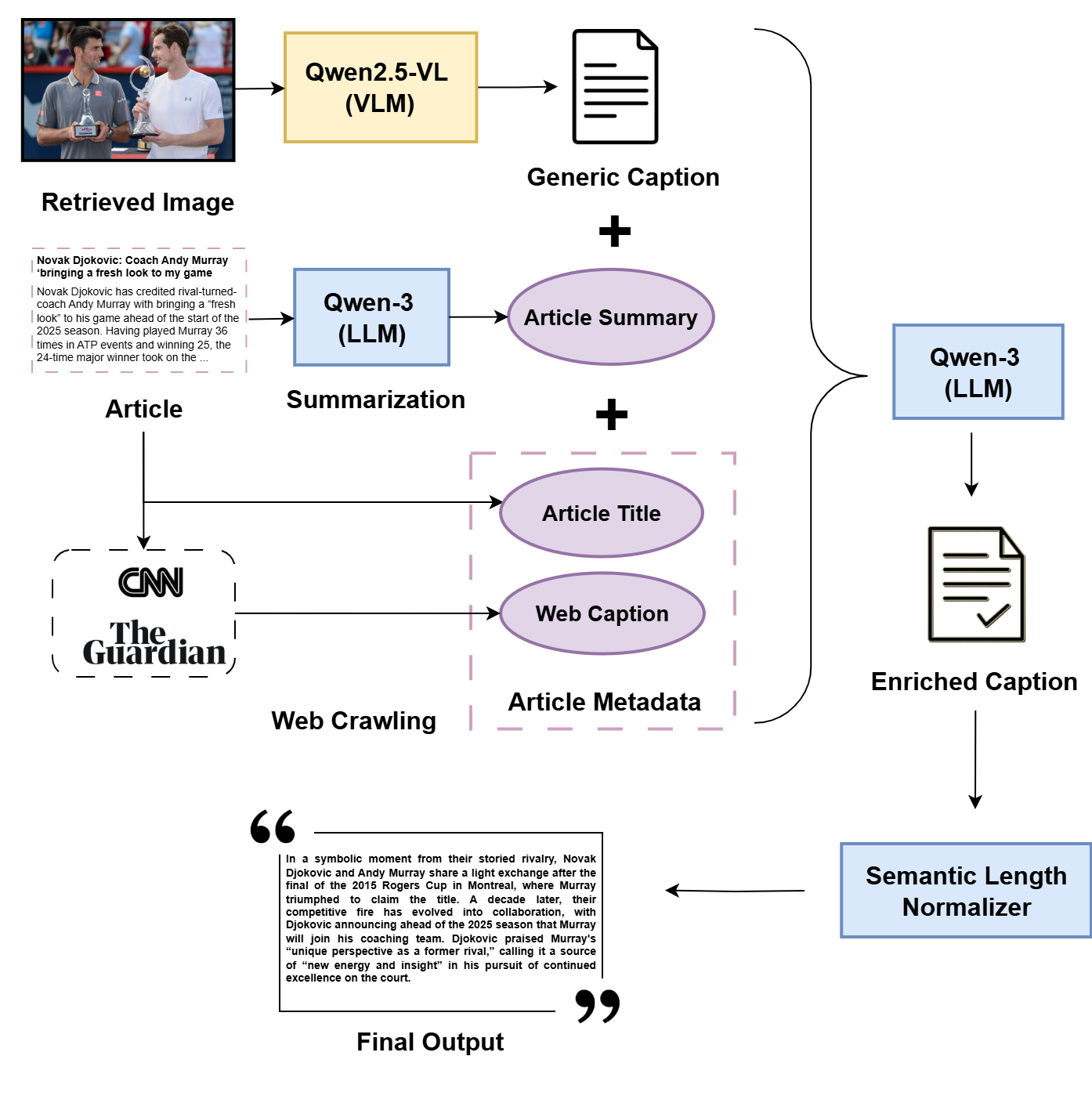}
    \vspace{-10mm}
    \caption{Caption enriching process: useful information (generic caption, article summary and web caption) is extracted to provide meaningful context before being fused to generate an event-enriched caption.}
    \label{fig:caption_enrich_pipeline}
    \vspace{-5mm}
\end{figure}

After retrieving the image and the corresponding article, we collect and extract multiple useful information from both the image and the article, before fusing them to generate an enriched caption. The process is shown in \autoref{fig:caption_enrich_pipeline}.

\textbf{Context Extracting:} We extract these useful information:
\begin{itemize}
    \item \textbf{Image's generic caption}: The image’s generic caption which provides intuitive descriptions of visible content, such as objects, scenes, and actions. This is generated using a vision-language model, specifically Qwen2.5-VL~\cite{qwen2.5-VL}, which processes raw pixel inputs to produce captions grounded solely in visual semantics, without any external context.
    \item \textbf{Image's web caption}: The caption displayed in the url of the article. Since the source of the articles are mainly from CNN or The Guardian, we build a simple web crawler combined with image-image matching to assign web caption each image. With the retrieved image and the corresponding article, we access the article through its url to get all available image-caption pairs. After that, we employ a vision encoder based on CLIP (Contrastive Language–Image Pretraining)~\cite{clip} to match the retrieved caption with the exact crawled image to get the corresponding web caption.
    \item \textbf{Article's summary:} The summarized article's content. To generate coherent and informative summaries of news articles, we employ large language model --- especially Qwen3~\cite{qwen3} with a simple prompt. 
\end{itemize}

\textbf{Caption Generating:} After gathering important and useful context, we employt the Large Language Model - Qwen3~\cite{qwen3} to fuse those context and generate the enriched caption. We apply an instruction-based prompting strategy to guide the model to create a caption which includes both visual information and key context from the article.

\subsection{Semantic Gaussian Normalizer}
\label{sec: Semantic Gaussian Normalizer}

\textbf{Motivation:}
One of the evaluation metrics for the challenge is the CIDEr score~\cite{vedantam2015cider} which evaluates the agreement between generated and reference captions:
\begin{equation}
\text{CIDEr-D}_n(c_i, S_i) = \frac{10}{m} \sum_j 
e^{ -\frac{(l(c_i) - l(s_{ij}))^2}{2\sigma^2} } \cdot 
\frac{ \min\left( \mathbf{g}^n(c_i), \mathbf{g}^n(s_{ij}) \right) \cdot \mathbf{g}^n(s_{ij}) }
{ \| \mathbf{g}^n(c_i) \| \| \mathbf{g}^n(s_{ij}) \| },
\end{equation}
where $c_i$ denotes the generated caption, $s_{ij}$ is a reference caption, $l(\cdot)$ represents the caption length, and $\mathbf{g}^n(\cdot)$ is the TF-IDF vector of n-grams.

This equation has the Gaussian penalty as a heavy penalty for captions which have even minor differences in caption length as compared to the reference caption. This penalty is indicated by the term $e^{ -\frac{(l(c_i) - l(s_{ij}))^2}{2\sigma^2} }$ which fines exponentially the caption length difference between the generated and the reference.

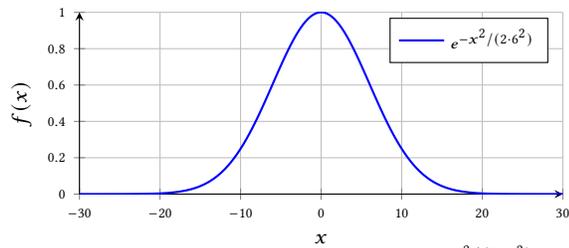
\begin{figure}[t!]
\centering
\begin{tikzpicture}
\begin{axis}[
    width=0.45\textwidth,      
    height=4cm,
    domain=-30:30,
    samples=200,
    xlabel={$x$},
    ylabel={$f(x)$},
    axis lines=left,
    grid=both,
    tick label style={font=\tiny},
    label style={font=\small},
    legend style={font=\tiny},
    legend pos=north east,
]
\addplot[blue, thick] {exp(-x^2 / (2 * 36))};
\addlegendentry{$e^{-x^2 / (2 \cdot 6^2)}$}
\end{axis}
\end{tikzpicture}
\vspace{-5mm}
\caption{The graph of the function $f(x) = e^{-x^2 / (2 \cdot \sigma^2)}$, where $\sigma = 6$ is normally used in CIDEr score. The variable $x$ indicates the caption length difference between the generated caption and the reference sentence; $f(x)$ indicates the penalty factor.}
\label{fig:gaussian-penalty-curve}
\vspace{-5mm}
\end{figure}

As illustrated in \autoref{fig:gaussian-penalty-curve}, even a 10-word difference in length can reduce the CIDEr score by over 75\%, highlighting the severity of the penalty.

\begin{figure}[!t]
    \centering
    \includegraphics[width=1\linewidth]{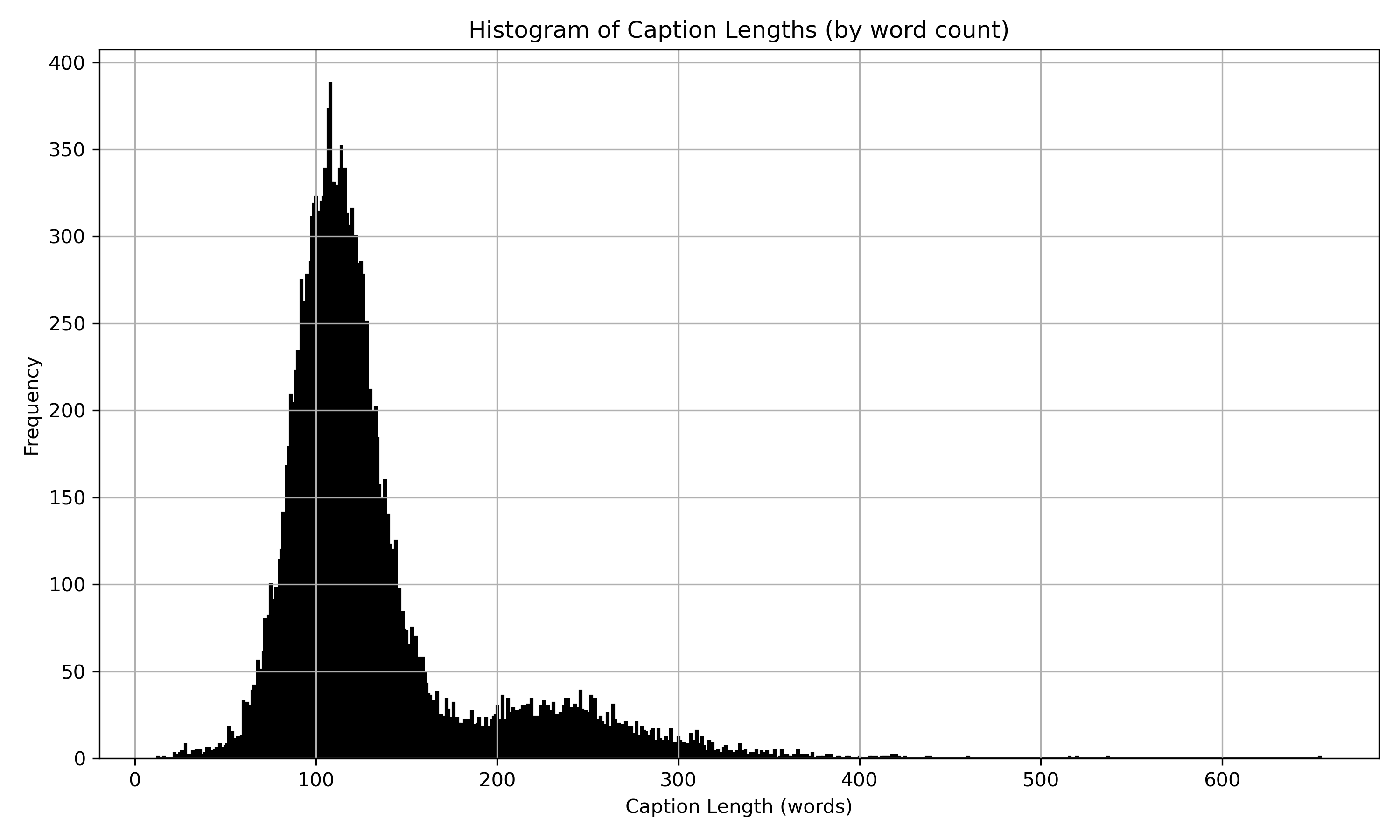}
    \vspace{-10mm}
    \caption{Generated caption word length distribution on training set.}
    \label{fig:wordCountTrain}
    \vspace{-5mm}
\end{figure}

For such motivation, we propose a trade-off strategy to manage caption length. If a generated caption exceeds a predetermined threshold, we truncate it. Conversely, if its length falls below this threshold, we enrich the caption with additional words. The primary objective is to align generated caption lengths closely with this optimal threshold. Empirically, we've observed that an ideal threshold lies between 90 and 120 words, as this range predominates the training set distribution (see \autoref{fig:wordCountTrain}, illustrating word counts in the training data). Given that CIDEr \cite{vedantam2015cider} assigns higher scores to rare words, such as named entities, our proposed algorithm also focuses on enhancing the relevance and informational content of captions, thereby improving alignment with evaluation metrics. This comprehensive strategy addresses both over-length and under-length caption generation challenges.

\textbf{Semantic Gaussian Normalizer} aims to enhance caption descriptions and mitigate the Gaussian length penalty inherent in CIDEr-D metrics \cite{vedantam2015cider} by semantically adjusting and normalizing generated caption lengths. This approach comprises three phases: {Gaussian Normalizer}, {Semantic Normalizer}, and {Entity Enricher}.

\textit{Gaussian Normalizer} truncates captions exceeding a predefined word threshold by simply cutting off any words beyond that limit. This method prioritizes adherence to length constraints.

\textit{Semantic Normalizer} also truncates over-length captions at a specific threshold. However, it intelligently preserves critical named entities within the caption. Instead of a blind cut-off, it removes less important, "normal" words first, ensuring that valuable named entities, which are often highly weighted by metrics like CIDEr, remain intact.

\textit{Entity Enricher} addresses under-length captions by automatically appending additional words to them until a desired minimum length is achieved. This enrichment specifically focuses on incorporating named entities drawn from various external text sources associated with the image, such as titles, article summaries, or web descriptions. This approach aims to make the caption more descriptive and to avoid length penalties from metrics like CIDEr-D that can penalize overly short captions.



\section{Experiments}

\subsection{Implementation Details}

For image and article retrieval, we used the DINOv2-Large model (1B parameters)~\cite{oquab2023dinov2} to extract global and patch-level features for image retrieval.

We observed from the reasoning ability of Qwen3~\cite{qwen3} that it could effectively recognize the relevance between the article content and the image generic caption. Hence, to enrich the caption with contexts, we added an example to make the model able to capture better information as follows:

\begin{tcolorbox}[colback=gray!5, colframe=gray!50!black, boxrule=0.4pt,
    sharp corners, enhanced, left=4pt, right=4pt, top=4pt, bottom=4pt,
    fontupper=\ttfamily\small, before skip=6pt, after skip=6pt]
\textbf{System Instruction:}  
You are an expert journalistic assistant tasked with creating detailed and informative image captions.  
Your goal is to combine information from image descriptions and the information of an article (from sources like The Guardian or CNN) into a single, enriched caption.  
The image sometimes seems irrelevant to the article, but try to use the context given by the article to generate an appropriate caption.  
Example: an article talking about a tennis event in the Olympics, but the photo is about the swimming event in the same Olympics. You need to take the scenario of the Olympics to generate the caption for the image relating to swimming.  
The output must be plain text without any formatting. The language of the caption MUST BE in English.  
The number of words of a caption should be from 100 to 140 words.

\vspace{0.5em}

\textbf{User Input: } 
Create a detailed caption by combining both the image caption and the information of the article I provide.  
Only return plain text with no formatting:

The image description:  
<generic\_caption>

The image caption from the article:  
<web\_caption>

The summary of the article:  
Title: <title>
<summary\_article\_content>
\end{tcolorbox}

Generation settings for both models are summarized in \autoref{tab:generation_settings}. Qwen3's thinking mode was enabled to enhance reasoning and contextual fusion, supported by a high token limit to prevent early truncation. Decoding parameters (temperature, top-p, top-k) were tuned for consistent and contextually aligned caption generation.

\begin{table}[t!]
\centering
\caption{Decoding settings and reasoning behavior for models used in the captioning pipeline.}
\vspace{-3mm}
\small
\setlength{\tabcolsep}{2pt} 
\begin{tabular}{p{2cm}ccccc}
\toprule
\textbf{Model} & \textbf{Temp.} & \textbf{Top-p} & \textbf{Top-k} & \textbf{Max Tokens} & \textbf{Thinking} \\
\midrule
Qwen2.5-VL-7B-Instruct
& $1\times10^{-6}$ & -- & -- & 10000 
& -- \\
Qwen3-14B
& 0.6 & 0.95 & 20 & 5000 
& True \\
\bottomrule
\end{tabular}
\label{tab:generation_settings}
\vspace{-3mm}
\end{table}

\begin{table*}[t!]
\centering
\caption{Ablation study results. RR, Qwen, GN, SN, and EE denote Re-ranking, Qwen2.5VL + Qwen3, Gaussian Normalizer, Semantic Normalizer, and Entity Enricher, respectively. All methods were built upon DINOv2~\cite{oquab2023dinov2} as the baseline.}
\vspace{-3mm}
\resizebox{\linewidth}{!}{
\begin{tabular}{ccccc|cccccc|cccccc}
\toprule
\multicolumn{5}{c|}{\textbf{Components}} & \multicolumn{6}{c|}{\textbf{Private Test Set}} & \multicolumn{6}{c}{\textbf{Public Test Set}} \\
\midrule
\textbf{RR} & \textbf{Qwen} & \textbf{GN} & \textbf{SN} & \textbf{EE} & \textbf{mAP} & \textbf{R@1} & \textbf{R@10} 
& \textbf{CLIPScore} & \textbf{CIDEr} & \textbf{Overall} & \textbf{mAP} & \textbf{R@1} & \textbf{R@10} 
& \textbf{CLIPScore} & \textbf{CIDEr} & \textbf{Overall}
\\
\midrule
 -- & -- & -- & -- & -- & 0.940 & 0.921 & 0.972 & -- & -- & -- & 0.987 & 0.980 & 0.996 & -- & -- & -- \\
\cmark & -- & -- & -- & -- & \textbf{0.982} & \textbf{0.977} & \textbf{0.988} & -- & -- & -- & \textbf{0.995} & \textbf{0.991} & \textbf{0.998} & -- & -- & -- \\
\midrule
 \cmark & \cmark & -- & -- & -- & 0.982 & 0.977 & 0.988 
& 0.870 & 0.145 & 0.44872 & 0.995 & 0.991 & 0.998
& 0.865 & 0.150 & 0.45779 \\
 \cmark & \cmark & \cmark & -- & -- & 0.982 & 0.977 & 0.988 
& 0.870 & 0.190 & 0.52527 & 0.995 & 0.991 & 0.998
& 0.865 & 0.256 & 0.61296 \\
 \cmark & \cmark & \cmark & \cmark & -- & 0.982 & 0.977 & 0.988 
& 0.870 & 0.194 & 0.53059 & 0.995 & 0.991 & 0.998
& 0.865 & \textbf{0.261} & \textbf{0.61904} \\
 \cmark & \cmark & \cmark & \cmark & \cmark & \textbf{0.982} & \textbf{0.977} & \textbf{0.988} 
& \textbf{0.870} & \textbf{0.205} & \textbf{0.54666} & \textbf{0.995} & \textbf{0.991} & \textbf{0.998} 
& \textbf{0.865} & 0.258 & 0.61619 \\
\bottomrule
\end{tabular}
}
\label{tab:CombinedResultsReordered}
\end{table*}

\subsection{Environmental Settings}

Our experiments were conducted on the test sets of the OpenEvents V1 dataset~\cite{openeventsv1}, provided as part of the EVENTA 2025 Grand Challenge \cite{eventa25}. 
We evaluated the methods using the official metrics defined by the EVENTA 2025 Challenge\footnote{\url{https://ltnghia.github.io/eventa/eventa-2025/track1}}.

\subsection{Ablation Study}

\subsubsection{Hyperparameter Search Using Public Test}

During the development phase, we performed an extensive hyperparameter search on the test set, specifically focusing on the maximum word threshold for our truncation strategies. We started with a threshold of 108 words, which was close to the observed peak in the training data's word count distribution (see \autoref{fig:wordCountTrain}). We then systematically explored nearby values through multiple submissions to evaluate their impact on the overall score. Our observations were as follows:

\begin{itemize}
    \item 108 words: Achieved an overall score of 0.54.
    \item 106 words: Achieved an overall score of 0.54.
    \item \textbf{104 words:} Achieved an overall score of \textbf{0.55}.
    \item 103 words: Achieved an overall score of 0.54.
\end{itemize}

These results indicate that a maximum word threshold of 104-105 words yielded the best performance on the test set for our algorithm. This iterative evaluation process provided valuable insights into the sensitivity of the CIDEr-D \cite{vedantam2015cider} score to caption length and allowed us to identify a highly optimal operating point for our proposed normalization strategies.

We notice that during the caption enrichment stage, configuring Qwen3 with a temperature of 0.85, top-p of 0.8, and top-k of 50 yields a suboptimal overall score of 0.52. This performance drop can be attributed to the increased randomness created by high temperature setting, which conflicts with the criterion of CIDEr---which rewards n-gram-level similarity. 

\subsubsection{Effectiveness of Components}

\begin{table}[t!]
\centering
\caption{Final results on the EVENTA 2025 private leaderboard of Track 1.}
\small
\setlength{\tabcolsep}{3pt}
\vspace{-3mm}
\begin{tabular}{cp{2.2cm}cccccc}
\toprule
\textbf{\#} & \textbf{Team Name} & \textbf{mAP} & \textbf{R@1} & \textbf{R@10} & \textbf{CLIP} & \textbf{CIDEr} & \textbf{Overall} \\
\midrule
1 & cerebro             & 0.991 & 0.989 & 0.995 & 0.826 & 0.210 & {0.55010} \\
2 & \textbf{SodaBread (Ours)} & \textbf{0.982} & \textbf{0.977} & \textbf{0.988} & \textbf{0.870} & \textbf{0.205} & \textbf{0.54666} \\
3 & Re: Zero Slavery    & 0.955 & 0.945 & 0.973 & 0.732 & 0.156 & {0.45148} \\
4 & ITxTK9              & 0.966 & 0.955 & 0.983 & 0.828 & 0.133 & {0.42002} \\
5 & noname\_            & 0.708 & 0.663 & 0.801 & 0.783 & 0.081 & {0.28241} \\
\bottomrule
\end{tabular}
\label{tab:leaderboard}
\vspace{-3mm}
\end{table}

\autoref{tab:CombinedResultsReordered} shows contributions of components in our proposed method. The re-ranking module with mutual nearest neighbor similarity applied to patch-level features significantly improves the retrieval scores, especially Recall@1 metric which has a great effect on the result of the captioning stage. Achieving a high Recall@1 score is crucial since it ensures the captioning process works on the right image and the right retrieved context. An inaccurately retrieved image may create an absolutely irrelevant caption, negatively affecting the CLIPScore and CIDEr score.

With the combination of generic caption from Qwen2.5VL and article context from Qwen3, the generated captions achieve the state-of-the-art CLIPScore. However, these captions create the moderate CIDEr score due to heavy Gaussian penalty as discussed in \autoref{sec: Semantic Gaussian Normalizer}.

On the other hand, our proposed Semantic Gaussian Normalizer, considerably increases the CIDEr score. The truncation step of captions using Gaussian Normalizer which purely cuts off words going beyond the threshold, contributes a notable increase. The Semantic Normalizer that truncates words based on their importance in CIDEr metric (typically avoid truncating named entities) contributes a greater increase in CIDEr score. 

Furthermore, Entity Enriching where collected named entities from the contexts are added into the caption successfully boosted the CIDEr score for the captioning task and directly increase the overall score.

\subsection{Final Results}

Our method achieved an overall score of 0.54666, ranking 2\textsuperscript{nd} in Track 1 of the EVENTA 2025 Grand Challenge~\cite{eventa25}. The top 5 results on the leaderboard are summarized in \autoref{tab:leaderboard}. These results highlight the effectiveness of our approach for the task of event-enriched image retrieval and captioning.


\section{Conclusion}
\label{sec5: Conclusion}

We introduce ReCap, a unified pipeline for event-enriched image retrieval and captioning that leverages DINOv2-based visual retrieval and large language models to generate context-aware captions grounded in real-world events. Our method integrates multiple context sources and incorporates a Semantic Gaussian Normalizer, a CIDEr-aware post-processing module designed to enhance caption quality. ReCap achieved strong performance in the EVENTA 2025 Grand Challenge, highlighting its effectiveness in both visual retrieval and event-enriched caption generation.


\begin{acks}
This research is supported by research funding from Faculty of Information Technology, University of Science, Vietnam National University - Ho Chi Minh City.
\end{acks}

\bibliographystyle{ACM-Reference-Format}
\balance
\bibliography{references.bib}

\end{document}